\documentclass{./styles/svproc}
\usepackage{graphicx}

\usepackage{url,hyperref,lineno,microtype,subcaption}
\usepackage{setspace}
\usepackage{amsmath}
\usepackage{algpseudocode}

\usepackage[ruled, vlined]{algorithm2e}
\SetAlCapNameFnt{\scriptsize}
\SetAlCapFnt{\scriptsize}
\SetAlFnt{\scriptsize}
\SetAlCapHSkip{0pt} 
\SetAlgorithmName{Algo}{Algo}{List of Algorithms}

\usepackage{array,tabularx}
\usepackage{xcolor}
\usepackage{placeins}

\usepackage{cite}

\usepackage[font=small,skip=0pt]{caption}
\newenvironment{conditions}
  {\par\vspace{\abovedisplayskip}\noindent
   \begin{tabular}{>{$}l<{$} @{} >{${}}c<{{}$} @{} l}}
  {\end{tabular}\par\vspace{\belowdisplayskip}}

\newenvironment{conditions*}
  {\par\vspace{\abovedisplayskip}\noindent
   \tabularx{\columnwidth}{>{$}l<{$} @{}>{${}}c<{{}$}@{} >{\raggedright\arraybackslash}X}}
  {\endtabularx\par\vspace{\belowdisplayskip}}

\begin{document}
\mainmatter              
%

\title{Fault-Tolerant Multi-Robot Coordination with Limited Sensing within Confined Environments}

\titlerunning{Fault-tolerant multi-robot coordination}  

\author{Kehinde O. Aina\inst{1} \and Hosain Bagheri\inst{2} \and
Daniel I. Goldman\inst{3}}
\authorrunning{K.O. Aina et al.} 
%
\tocauthor{Kehinde O. Aina, Hosain Bagheri, Daniel I. Goldman}
\institute{Institute for Robotics and Intelligent Machines, Georgia Institute of Technology, USA. \\ 
\and
School of Biological Sciences, Georgia Institute of Technology, USA. \\ 
\and
School of Physics, Georgia Institute of Technology, USA. \\ 
\email{\{\textsuperscript{1}kaina3, \textsuperscript{2}hbagheri, \textsuperscript{3}daniel.goldman\}@gatech.edu} }

\maketitle              

\begin{abstract}
As robots are increasingly deployed to collaborate on tasks within shared workspaces and resources, the failure of an individual robot can critically affect the group's performance. This issue is particularly challenging when robots lack global information or direct communication, relying instead on social interaction for coordination and to complete their tasks. In this study, we propose a novel fault-tolerance technique leveraging physical contact interactions in multi-robot system, specifically under conditions of limited sensing and spatial confinement. We introduce the ``Active Contact Response'' (ACR) method, where each robot modulates its behavior based on the likelihood of encountering an inoperative (faulty) robot. Active robots are capable of collectively repositioning stationary and faulty peers to reduce obstructions and maintain optimal group functionality. We implement our algorithm in a team of autonomous robots, equipped with contact-sensing and collision-tolerance capabilities, tasked with collectively excavating cohesive model pellets. 
Experimental results indicate that the ACR method significantly improves the system's recovery time from robot failures, enabling continued collective excavation with minimal performance degradation. Thus, this work demonstrates the potential of leveraging local, social, and physical interactions to enhance fault tolerance and coordination in multi-robot systems operating in constrained and extreme environments. 

\end{abstract}
\section{Introduction}
%
Multi-agent systems are preferred over single-agent systems in domains that require distributed execution or control, collaboration, and concurrency ~\cite{lesser1999multi, chen2009integrating, drew2021multi, ROUSSET201627}. The key to successful multi-agent coordination lies in the system's robustness and fault tolerance, which are essential for maintaining continuous operation and achieving the group's task objectives. While robustness ensures the system performs reliably under uncertainties and disturbances, fault tolerance enables it to continue functioning effectively despite the failure of one or more agents. Swarm robotics, a robust multi-agent system, has been applied in various fields such as environmental monitoring 
~\cite{lesser1999multi, valckenaers2007applications}, military demining ~\cite{baudoin2010using}, and healthcare ~\cite{shakshuki2015multi}. However, the inherent uncertainty and non-stationarity in swarm systems can lead to unforeseen failures, jeopardizing the mission. 

Swarm robotics draws inspiration from the principles of coordinated interaction seen in self-propelled entities that exhibit collective behaviors 
~\cite{sidd_inproceedings, article_Schmickl, Slavkov2018, article_Siddharth_and_Wilson, 8794124}. For instance, social insects demonstrate selective engagement in densely populated environments lacking global cues ~\cite{Gravish2012, gordon1996organization, osti_10427980}. Considering the complexity of their environments, physical and localized interactions enable their robustness and fault tolerance, allowing for rapid recovery in the face of disasters ~\cite{monaenkova2015behavioral, Gravish2015, Gravish9746}. However, unlike living systems, conventional robots typically lack mechanical compliance. Therefore, most robotic collective control strategies focus on avoiding physical interactions to prevent collisions ~\cite{Souliman2014, HernandezMartinez2011, 9123564, Arslan2016}. On the other hand, multi-robot or swarm control strategies necessitating a high level of coordination and control (e.g., ~\cite{Roy2019}) are often impractical and can lead to catastrophic failures in environment where precision and accuracy are important ~\cite{TAYLOR2021103754, BORRMANN201568}.

\begin{figure}[t!]
\begin{center}
\includegraphics[width=11cm]{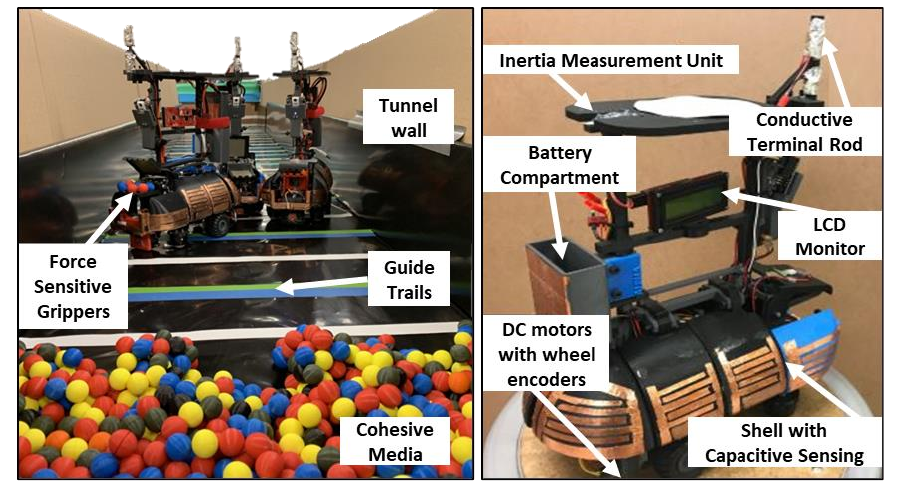} 
\end{center}
\caption{A model constrained and confined environment for multi-robot collective excavation of cohesive granular media. Robots respond to environmental cues using onboard sensors. Robot length is 32 cm.}
\label{fig:robot_setup} 
\end{figure}

Motivated by these challenges, we explore fault tolerance in autonomous robots equipped with limited sensing and operating in uncertain environments. Our robot models are simplified but include core functionalities applicable to real-world systems like warehouse or domestic robots. Consider the swarm robotic scenario depicted in Figure~\ref{fig:robot_setup}, where wheel-driven ellipsoidal shape robots perform tasks in a crowded and confined environment without centralized control or direct communication. These robots 
rely on local cues and simple onboard sensors, such as capacitive touch sensors and inertial measurement units (IMUs). This sets them apart from other systems that rely on overhead tracking or positioning systems for autonomous navigation and control. This aligns with certain examples of swarm robotics systems where robots must participate in local and social or contact interactions to collectively achieve their tasks ~\cite{Siddharth_1109, frontiers_kaina}.
In this context, a malfunctioning robot in a critical area can cause cascading effects of difficult-to-resolve interference and congestion. This can drastically hinder task completion and may result in severed pathways that disrupts the workflow ~\cite{science_paper}.

Our previous work developed principles that leverage local and physical interactions within dense active systems to address the challenges of robot coordination in crowded environments ~\cite{science_paper, frontiers_kaina}. Physical constraints imposed by such an environment make collisions and jamming unavoidable. We observe how adaptive idleness produces emergent coordinated flow and enhances the performance of robotic collectives by generating workload inequality and adapting reversal behaviors ~\cite{science_paper, frontiers_kaina}. These behaviors collectively emerge through contact interactions, enabling robots to adapt to the dynamic structure of the environment. 

Building on this, we introduce a fault-tolerance technique that improves a multi-robot system's ability to recover from individual robot failures. In particular, we propose an egocentric dynamic contact map that enables robots to localize and retain their contact interaction history. Upon encountering another robot, a robot can use this map to predict whether it has contacted a faulty stationary robot. It then either attempts to actively displace the stationary robot to a less obstructed configuration, thereby facilitating its eventual exit from the tunnel (Active Contact Response), or avoids further interaction to prevent it from being pushed further into the tunnel and causing additional obstruction (Passive Contact Response). This leads to emergent behavior where the group collectively repositions the faulty robot, minimizing the impact on overall performance. 

\section{LITERATURE REVIEW}
In the past few decades, researchers have explored several approaches to improving fault tolerance in multi-agent systems. Failures can arise from software malfunctions, hardware issues, or a combination of both ~\cite{5229934}. While software-related failures are often easier to detect and resolve due to the flexibility of provisioning and cloning software systems to create redundancy,  fault tolerance in physical systems presents more significant challenges. This is because physical systems often lack the flexibility to create redundancy schemes needed to easily resolve or substitute faulty components. 

To enhance fault tolerance in multi-agent systems, various redundancy schemes have been proposed, including agent replication and high-availability hybrid architectures ~\cite{5229934, inproceedings_Almeida, Kumar_Sanjeev}. Fedoruk et al. ~\cite{fedoruk2002improving}, along with other researchers, ~\cite{Marin2001TowardsAF, guessoum2005adaptive, 1639670, 5952430} 
investigated replication as a method to maintain copies of agent states across systems. This approach is effective in software-based systems, where faulty agents can be detected and substituted with minimal disruption. However, in physical systems where mechanical compliance is lacking, the problem extends beyond replicating agent states to ensuring seamless recovery from hardware failures.

Fault detection in multi-agent systems is often achieved through inter-agent communication via ``keep alive'' messages or periodic status exchanges ~\cite{Faci_fault_tolerant_mas, FaultToleranceMAS}. While communication is an essential part of detecting and maintaining fault tolerance in multi-agent systems, it introduces complexity and load factors in that it often requires the integration of information exchange channels among numerous agents, 
potentially leading to decreased system performance.

Several researchers have studied fault tolerance specifically in multi-robot systems. For instance, Park et al. ~\cite{7487153, 7353788} proposed a decentralized control law to address the multi-robot rendezvous problem, consisting of nonconforming or faulty robots. The technique is based on distributed control policy that is robust to undetected faults in the system. Al Hage et al. ~\cite{7535869} developed a fault-detection method for the collaborative localization task of multi-robot. The approach relies on the Informational form of the Kalman filter to detect and exclude faulty sensors from the team. Yang et al. ~\cite{8854221} outlined several approaches for fault tolerance in cooperative control of multi-robot systems. In multi-agent systems, increased redundancy and flexibility bring about heightened complexity and challenges. Hence, the authors introduced four categories of fault-tolerant cooperative control methodologies: individual, cooperative, topology reconfiguration-based, and composition reconfiguration-based.

Our approach differs from traditional fault tolerance techniques as it does not rely on explicit fault detection and identification. Instead, we employ a passive fault-tolerance scheme in which robots leverage local interactions to collectively address failures. This method is particularly effective in noisy and partially observable environments, where explicit fault identification may be impractical. 
Robots make decisions based on their private contact interaction history, potentially providing scalability to our method. Furthermore, the decentralized nature of the ACR method enables the collective to accomplish the task without requiring the activity history of neighboring robots. This presents a distinct advantage over numerous established swarm robotics methods, which rely on shared information between agents to complete a collective task ~\cite{article_Schmickl, article_Siddharth_and_Wilson, 8794124}. 

\section{MULTI-ROBOT COLLECTIVE EXCAVATION}

The multi-robot collective excavation problem, as described in previous work ~\cite{frontiers_kaina, aguilar2018collective}, involves a team of homogeneous robots assigned to continuously retrieve cohesive model granular media (``pellets'', 3D-printed hollow spheres with embedded loose magnets) within a narrow and confined tunnel (Figure~\ref{fig:robot_setup}). The objective is to maximize the number of pellets collected within a fixed time frame. Each robot is equipped with basic sensors such as an IMU, wheel encoders, capacitive touch sensors, force-sensing resistors, magnetometers, and terminal rod capacitive sensors, but lacks global control and access to other robots' states.

Each robot begins a trip by departing from the home area and following guidance trails towards the excavation site, which is detected using a magnetometer at the robot’s anterior base. Upon locating the cohesive pellets via magnetic field sensing, the robot initiates the excavation routine in an attempt to collect the pellets. If successful, the robot transports the pellets back to the home area. While navigating through the tunnel, robots may detect and differentiate contacts or collisions with other robots or the tunnel walls. When robot-to-robot contact is detected during transit to the excavation area, a robot may abandon or ``give up'' its digging attempt based on a predefined reversal probability, $P_r$ (Figure~\ref{fig:fsm_diagram}), a behavior referred to as \emph{Reversal Behavior} ~\cite{frontiers_kaina}. This behavior is also triggered when a robot is unable to reach the excavation site within a specified time frame, $T_g$, preventing prolonged tunnel congestion when the robots are in a gridlock. 

\subsection{Robot Controller}
\begin{figure}[t!]
\begin{center} 
\includegraphics[width=12.2cm]{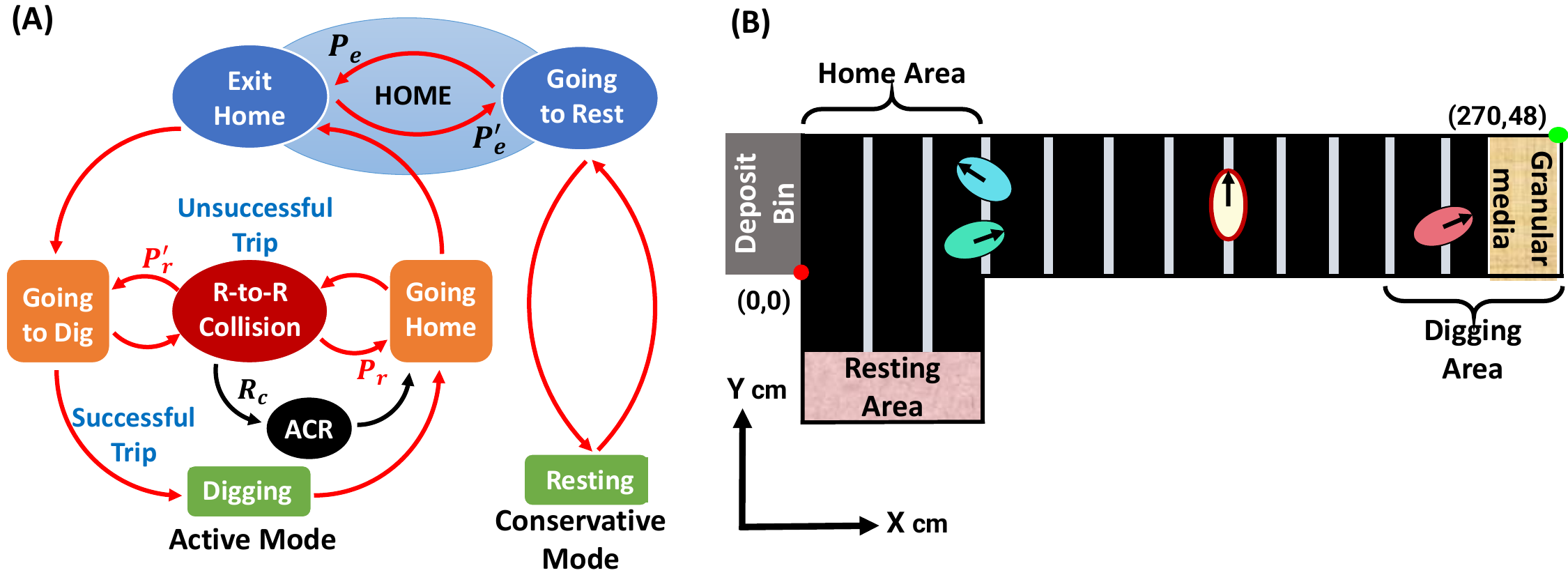}  
\end{center}
\caption{\textbf{(A)} 
Block diagram of the individual robot controller based on Finite State Machine. $P_e$ is the probability of switching to Active Mode. $P_e'$ is the probability of switching to Conservative Mode. $P_r$ is the probability of ``giving up'' upon detecting contact with another robot. $R_c$ is the probability that the contact is from a faulty robot. \textbf{(B)} Schematic of experimental setup showing three active robots (green, blue, and red), and one faulty robot (yellow). The origin of coordinate system is indicated.}

\label{fig:fsm_diagram}
\end{figure}

We adopt a behavior-based control model consisting of two layers of a Finite State Machine (FSM) \cite{PANESCU2007121}. The top layer uses a probabilistic FSM model, in which the robot decides which operational mode to enter, determined by an entrance probability parameter, $P_e$. The robots choose whether to operate in Active Mode, meaning they enter the tunnel to participate in pellets excavation, or in Conservative Mode, where they remain in the home area and refrain from participating in excavation. If a collision occurs while entering the tunnel, the robot decides to ``give up'' and return home based on the reversal probability $P_r$. 
At the bottom layer, the robot initiates state transitions in response to physical cues from the environment. Each of the sensors on the robot (Figure~\ref{fig:robot_setup}) is configured with a trigger condition that enables it to transition between different states. Figure~\ref{fig:fsm_diagram} shows the model of the robot's controller. The robots can exist in one of several internal states depending on their operational mode and environment interactions. These internal states include: 

\begin{enumerate}
    \item \emph{Going to Dig}: This state is activated at the beginning of a trip, $k$, if the robot decides to enter the tunnel for excavation based on the probability $P_e\left(k\right)$. It is the first state in the robot's Active Mode. 
    \item \emph{Digging}: Once the robot approaches the cohesive granular media, the magnetometer at the robot’s anterior base detects the magnetic field, prompting the robot to start its excavation routine. 
    \item \emph{Going Home}: This state captures the robot's return back home, either following a successful pellet retrieval (Successful Trip) or an unsuccessful pellet retrieval (Unsuccessful Trip). The controller guides the robot from the tunnel to the home or deposit area. The entrance probability for the next trip, $P_e(k+1)$, is updated based on the success of pellet retrieval (Algorithm \ref{acr_algorithm}). 
    \item \emph{Resting}: A robot transitions into this state when it chooses not to enter the tunnel, triggering the Conservative Mode. A robot will enter this state at the beginning of a trip based on resting probability $P_e'\left(k\right) = 1 - P_e\left(k\right)$. The robot remains idle in the home area for a fixed amount of time, $T_r$, avoiding tunnel traffic and reducing congestion. 
    \item \emph{Collision}: This state is triggered when a robot makes contact with the tunnel wall or another robot through its capacitive sensors. The robot can respond in one of three ways depending on its previous state:
    \begin{enumerate}
        \item{Default Maneuver}: The robot executes a random turning maneuver to resolve the collision and continue with its task. 
        \item{Reversal Behavior}: The robot abandons its task to avoid further congestion and returns to the home area.
        \item{Active Pushing}: The robot attempts to actively push the obstructing robot or obstacle to reposition it into a less obstructive configuration.
    \end{enumerate}
    Note that a robot can transition to collision state from any state other than the Resting state. This is omitted in Figure~\ref{fig:fsm_diagram} $\boldsymbol{(A)}$ to avoid visual clutter. \\
\end{enumerate}

\vspace{-0.5cm}

We developed a stochastic model with two parameters to control the entrance rate and reversal rate (give-up rate) of the robots. Let $P_e\left(k\right)$ be the tunnel entrance probability and $P_r\left(k\right)$ be the reversal probability of each robot at trip attempt number $k$. A trip begins when a robot samples from the entrance probability, $P_e\left(k\right)$ and decides whether to ``go in and dig'' or ``stay at home and rest''. This parameter controls the number of robots in the tunnel, directly influencing tunnel density and congestion rate. The reversal probability, $P_r\left(k\right)$, on the other hand controls how a robot responds to a collision when it occurs, directly influencing the duration of congestion in the tunnel. A robot samples from this parameter and decides if it should ``give up'' or to continue its journey.

\section{ACTIVE CONTACT RESPONSE METHOD}
Here we introduce the Active Contact Response (ACR) algorithm, which maintains a transient space-time contact map of each robot's contact encounters in the tunnel. This enables robots to infer the likelihood of contact with a faulty robot and take the necessary actions to mitigate persistent interference. Initially, each robot's contact map is initialized to zero, implying that they have no prior information about the distribution of active and passive (faulty) robots in the environment. 

As collective excavation progresses, each robot continuously updates its contact map based on the contact type (e.g., robot or wall) it encounters. Using the capacitive touch sensors (Figure \ref{fig:robot_setup}), robots can distinguish between robot-to-wall and robot-to-robot contacts.  
Adopting a time-varying cumulative map to represent interaction history enables the robots to develop a belief state regarding the potential location of a faulty robot in the environment, should one exist. Specifically, the robot updates: 1) the frequency of each contact type, $c_r$ and $c_w$, and 2) the longitudinal position, $l$, where the contact occurred. The contact map, 
$M(.,.)$, which is a function of contact type and longitudinal position, undergoes conditional updates using the following equation (Algorithm \ref{map_update_algo}):

\vspace{-0.2cm}

\begin{equation} 
  \begin{aligned}
    M(c_r,l)_{t+1} = M(c_r,l)_t + \omega_r*W \\
    M(c_w,l)_{t+1} = M(c_w,l)_t + \omega_w*W \\
 \end{aligned}
\end{equation}

\vspace{-0.3cm}

where:
\vspace{-0.5cm}
\begin{conditions}
   \qquad c_r & = & robot-robot contact \\
   \qquad c_w & = & robot-wall contact \\
   \qquad l & = & estimated contact position \\
   
   \qquad M(c_r,l) & = & frequency of robot-robot contact at position $l$ \\
   \qquad M(c_w,l) & = & frequency of robot-wall contact at position $l$ \\
   
   \qquad \omega_r & = & conditional likelihood of detecting a robot-robot contact \\
   \qquad \omega_w & = & conditional likelihood of detecting a robot-wall contact \\
   \qquad W & = & weighting factor \\
\end{conditions}

\vspace{-0.2cm}

The conditional likelihood values $\omega_{r}$ and $\omega_{w}$ are obtained through the calibration of a contact sensor's sensitivity to robot and wall contacts respectively. 
 At regular intervals, the frequency of each contact type is decreased by a constant, $\beta$, to reduce the effect of noise and the non-stationarity of the multi-robot environment: $ M(c,l)_{t+1} = M(c,l)_t - \beta $.


Each robot uses its dynamically updated contact map to make Active Contact Response decisions based on the likelihood $R_c$ that a robot-to-robot contact is originating from a faulty robot. The likelihood is calculated as follows: 

\vspace{-0.2cm}

\begin{equation} 
  \begin{aligned}
        R_c = \frac{\qquad e^{\sum M(c_r,l)}}{\qquad e^{\sum M(c_r,l)} + e^{\sum M(c_w,l)}} \\
 \end{aligned}
\end{equation}

\vspace{-0.1cm}

Where $\sum $ is over the projected length of the robot.
The rationale for this expression relies on the normal distribution of robot contact frequency along the tunnel's length, where the mean contact frequency correlates with the location of the faulty robot. See algorithms \ref{acr_algorithm} and \ref{map_update_algo} for the implementation pseudo code.

An active robot going into the tunnel to excavate seeks to resolve collisions and improve group performance by voluntarily ``giving up''. However, a wrongly positioned inactive or faulty robot will impede group performance by causing persistent jamming or gridlock. As mentioned in the previous section, if a contact is detected, a robot has three options, conditioned on the internal state: 1) execute the default maneuver to resolve contact - passive contact response, 2) execute reversal behavior to avoid tunnel clogging - reversal behavior, or 3) execute the ACR maneuver to displace the supposed faulty robot to a less obstructive configuration. For instance, when a robot has a high confidence ($R_c\approx1$) that it has encountered a stationary faulty robot en route to the excavation site, it maneuvers around it by implementing either the reversal behavior or the passive contact response. However, upon the robot returning home from excavation, it 

\begin{minipage}{0.51\textwidth}
\begin{algorithm}[H] 
    \scriptsize
    \caption{Active Contact Response} 
    \label{acr_algorithm}
 Initialize: $k=1$, $P_e(k)=1$, $P_r=0.64$, $t_{prev}=0$, $M(:,:)=0$\;
 Set: expt. duration,$T_e$; prob. update,$\Delta r$; ``giving up'' time,$T_g$; resting time,$T_r$; update rate,$\Delta C$; R-R likelihood,$\omega_r$; R-W likelihood,$\omega_w$; weight,$W$; decay rate,$\beta$.
 
 \While{$t < T_e$}{ 
    \eIf{$p \sim U(0, 1) < P_e(k)$}{ 
        should\_dig = 1 \\
        \While{should\_dig == 1}{  
            \If{ContactDetected}{ 
                UpdateContactMap() \\                
                \If{$q \sim U(0, 1) > P_r(k)$}{ 
                should\_dig = 0 \\
                should\_go\_home = 1 \\
                break 
                }
            }
            \If{GrabsPellets $\boldsymbol{or}$ (t $>$ $T_g$)}{ 
                should\_dig = 0 \\
                should\_go\_home = 1 \\
                break 
            }
            \If{$(t - t_{prev}) > \Delta C$}{ 
                \text{for $c=0$; $c<C$; $c{+}{+}$} \\
                 \quad \text{for $l=0$; $l<L$; $l{+}{+}$} \\
        		\quad \quad $M(c,l) = M(c,l) - \beta$  \\
                $t_{prev} = t$ 
	       }
        } 
        
        \While{should\_go\_home == 1}{  
            \If{ContactDetected}{ 
                UpdateContactMap() \\
            }
            \eIf{ReturnsHomeWithPellets}{ 
                 $P_{e}\left(k+1\right) = \max(1.0, P_e\left(k\right) + \Delta r)$ \\
                 should\_go\_home = 0 \\
                 break 
            }{
                $P_{e}\left(k+1\right) = \max(0, P_e\left(k\right) - \Delta r)$ \\
                 should\_go\_home = 0 \\
                 break 
            }            
            \If{$(t - t_{prev}) > \Delta C$}{ 
                \text{for $c=0$; $c<C$; $c{+}{+}$} \\
                \quad \text{for $l=0$; $l<L$; $l{+}{+}$}  \\
                \quad \quad $M(c,l) = M(c,l) - \beta$ \\
                $t_{prev} = t$ 
        	}
        }        
    }{
        \While{$(t-t_{cur}) < T_r$}{ 
            RestRobot() \\
       }
    }
    $k = k + 1$
 }
\end{algorithm}
\end{minipage}
\hfill 
\begin{minipage}{0.44\textwidth}
\begin{algorithm}[H]
    \caption{UpdateContactMap()}
    \label{map_update_algo}
    \scriptsize
        $l = getRobotXCoordinate()$
        
        \eIf{robot-robot contact}{  
           $M(c_r,l)_{t+1} = M(c_r,l)_t + \omega_r*W$ \\
           $M(c_w,l)_{t+1} = M(c_w,l)_t + (1-\omega_r)*W$ \\
        } {
           $M(c_r,l)_{t+1} = M(c_r,l)_t + (1-\omega_w)*W$ \\
           $M(c_w,l)_{t+1} = M(c_w,l)_t + \omega_w*W$ \\
        }
        \[ R_c = \frac{\qquad e^{\sum M(c_r,l)}}{\qquad e^{\sum M(c_r,l)} + e^{\sum M(c_w,l)}}\]  \\
        \eIf{$State = GoToDig()$}{ 
            \eIf{$q \sim U(0, 1) > P_r$}{ 
                GoToHome() \\        
             } {
                PassiveContactResponse() \\
             }
        } {
            \eIf{$q \sim U(0, 1) < R_c $}{ 
                ActiveContactResponse() \\
             } {
                PassiveContactResponse() \\
             }
        }
\end{algorithm}

\vspace{0.5cm}
 will bias its behavior by continuously pushing the stationary faulty robot until it can proceed without obstruction (i.e., ACR maneuver). 
  However, if the robot fails to make progress in repositioning the faulty robot within a set time frame $T_g$, it transitions to passive contact maneuver. In passive contact maneuver, robot ceases biased behavior and resorts to default contact resolution maneuvers to avoid further congestion and gridlock. This fallback mechanism adds robustness to the ACR method, ensuring that robots do not become "stuck" in repetitive, ineffective maneuvers.
\end{minipage}

\section{EXPERIMENT}

We implemented the Active Contact Response (ACR) algorithm described in Section IV on physical multi-robot system to evaluate its performance against our baseline algorithm. In the baseline algorithm, robots neither employ a contact map nor actively respond to or bias their behavior when they detect collision with other robots. Instead, robots either attempt to resolve collision and continue (Passive Contact Response), or ``give up'' on excavation efforts (Reversal behavior). 

In our experimental setup, three active robots are tasked with excavating pellets, while a fourth stationary robot, powered off, serves as a model for the faulty robot (Figure~\ref{fig:fsm_diagram}). The active robots were unaware of the presence of a ``faulty'' robot. Instead, they had to infer the likelihood of encountering a stationary robot based on the distribution of collision frequencies along the tunnel length. We designed the tunnel width to accommodate two robots either moving in the same direction, or moving in opposing parallel directions.  

At the start of each experiment, we placed the faulty (stationary) robot perpendicularly in the middle of the tunnel (Figure~\ref{fig:fsm_diagram}). The perpendicular orientation of the faulty robot with respect to the length of the tunnel was selected to ensure sufficient contact interaction with the active robots. Positioning the faulty robot in the middle of the excavation path ensures it remains adequately distant from both the home area and the excavation site.

It is important to note that the position of the faulty robot is critical in determining the group's overall performance. For example, the interference is maximum when the passive robot's orientation is perpendicular to the longitudinal direction of the tunnel, as well as when its position is close to the excavation area, where robots would need to execute turning maneuvers more frequently (Figure~\ref{fig:fsm_diagram}). This faulty robot configuration further constrains the dimension of the excavation arena and impacts the group's excavation performance.

On the other hand, interference is minimal when the faulty robot's orientation is parallel to the longitudinal direction of the tunnel and its position is close to the home area, where there is enough space for the robots to make turns without much collision or interference. This configuration is termed less obtrusive because the robots can still complete their excavation and deposition tasks.

The ACR algorithm functions by implicitly minimizing interference caused by the faulty robot through indirect displacement and reorientation into a less obstructive configuration. As described in the previous section, the active robots use their cumulative sum of contact experiences to estimate the likelihood of encountering a stationary robot. In the case when there is no faulty robot in the tunnel, the collision map or collision frequency across the tunnel length would follow a uniform distribution. This implicitly reduces the probability that a contact interaction stems from a stationary robot.

 The experiment was conducted in a tunnel of fixed length, with each robot using independent FSM (Section 3). The experiment duration was set to 30 minutes, with the initial tunnel entrance probability, $P_e$, set to 1, ensuring that all robots started in Active Mode. Key hyper-parameters, including reversal probability ($P_r$), resting time ($T_r$), and give-up time ($T_g$), were optimized through preliminary trials. Adjustments to these parameters may lead to varied learning and convergence times, but the robots will demonstrate similar overall behavior. During the experiment, $P_e$ was gradually adjusted based on the robot’s excavation success, as described in the workload distribution methods in our prior work \cite{frontiers_kaina}.

At the conclusion of each experiment, the final configuration (position and orientation) of the faulty robot was measured to assess the level of obstruction it introduced to the system. Three experimental trials were conducted for both the ACR and baseline methods to ensure consistency and reliability in the results. Each trial provided valuable insights into how the positioning and interaction of the faulty robot impacted overall excavation performance.

\vspace{-0.3cm}

\begin{figure}[t!]
\begin{center}
 \includegraphics[width=11cm]{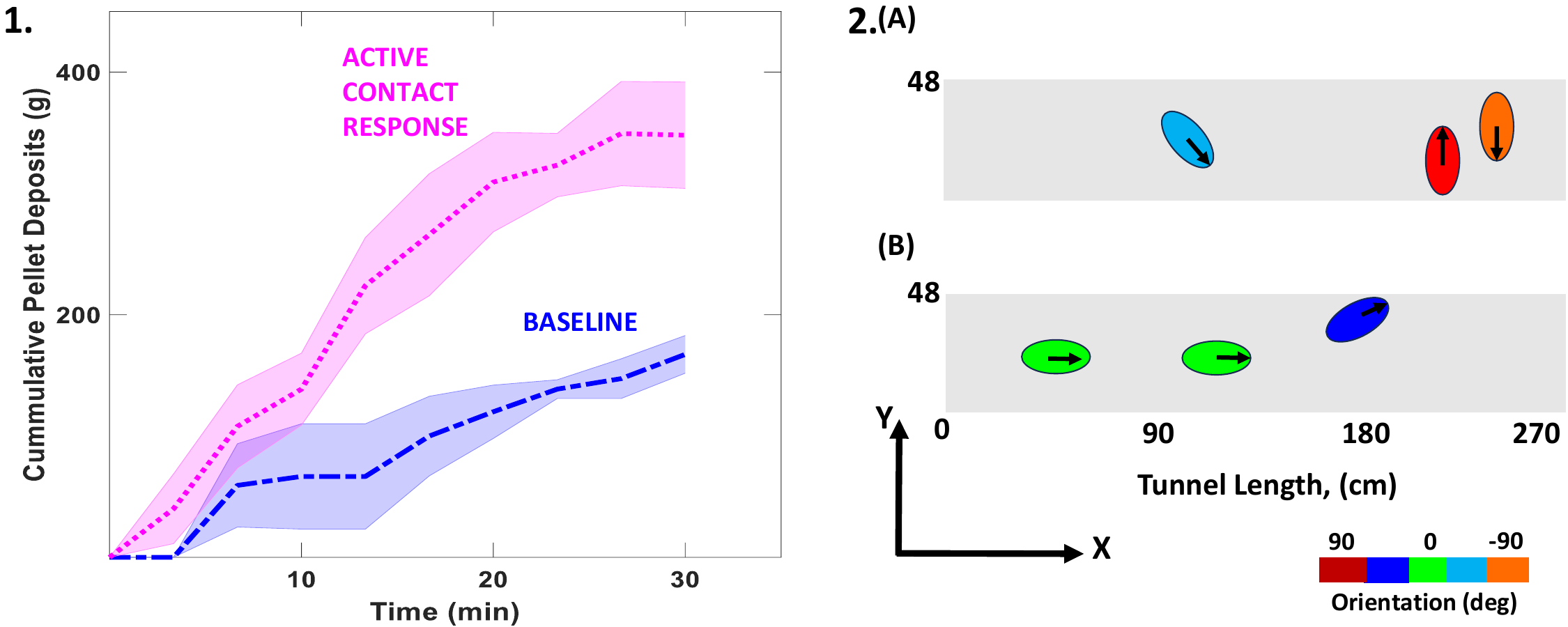} 
\end{center}
\caption{\textbf{1.} Comparison of excavation performance between the Baseline and ACR methods across three trials. Solid-dotted and dash-dotted lines represent the mean excavated pellets, with shaded areas indicating the standard deviation. \textbf{2.} Final configuration of the faulty robot after each trial: \textbf{(A)} Three Baseline trial, and \textbf{(B)} Three ACR trials. .
}
\label{fig:excav_perf}
\end{figure}

\begin{figure}[h!]
\begin{center}
\includegraphics[width=12.2cm]{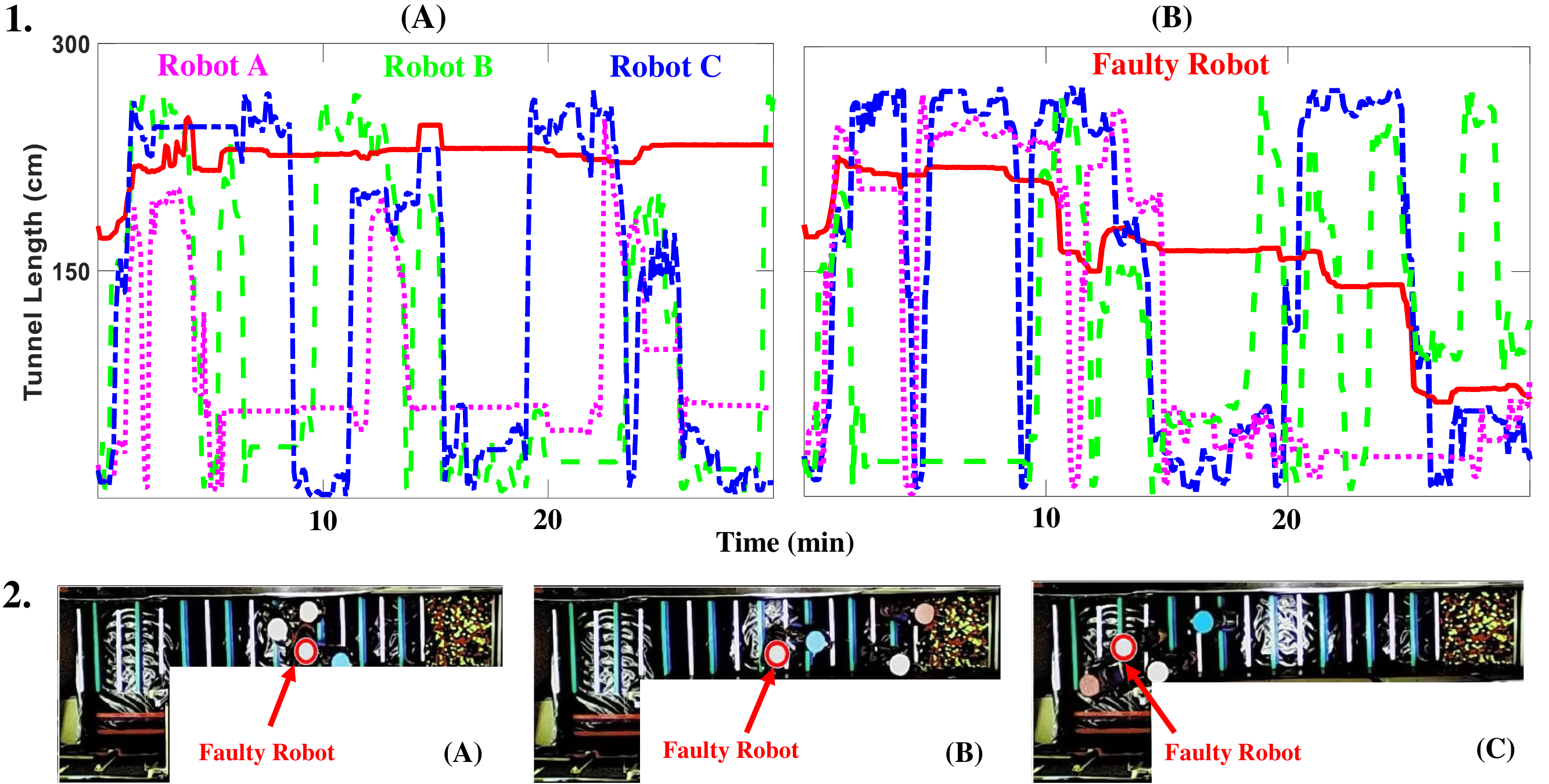}
\end{center}
\caption{\textbf{1.} Robot trajectories tracked overtime for \textbf{(A)} Baseline method, and \textbf{(B)} Active Contact Response (ACR) method. Tunnel spans from 0 cm (home area) to 300 cm (digging area). \textbf{2.} Top-view snapshots of the ACR experiment: \textbf{(A)} Initial configuration of faulty robot obstructs traffic flow, resulting in clustering at the middle of the tunnel. \textbf{(B)} Subsequently, active robots update their collision map and maneuver around the faulty robot to reach the excavation area. \textbf{(C)} Finally, active robots effectively reposition faulty robot into a less obstructive configuration through active pushing and nudging.}
\label{fig:robot_trajectory}
\end{figure}

\vspace{-1cm}

\section{RESULTS AND DISCUSSION}

Figure~\ref{fig:excav_perf}.1 highlights the cumulative pellet deposition for both the ACR and Baseline methods (measured by monitoring of a scale onto which pellets were deposited).  
Initially, both methods demonstrate similar excavation rates during the first 10 minutes, but a divergence occurs as the experiment progresses. The Baseline method exhibits a consistently lower excavation rate throughout the remainder of the experiment. This decline is largely due to frequent collisions with the faulty robot, which led to more robots giving up and resting rather than attempting to reposition the faulty robot to a less obstructive configuration. In contrast, the ACR method more effectively managed the presence of the faulty robot by actively repositioning it, resulting in a significant improvement in the overall excavation rate. By the end of the 30-minute experiment, the ACR method had excavated and deposited nearly twice as many pellets as the baseline method.

 At the conclusion of the experiment, we measured the final position and orientation of the faulty robot to assess obstruction levels. Optimal configurations are characterized by orientations close to zero degrees (parallel to the tunnel) and positions near the home area, which minimize obstruction. As shown in Figure~\ref{fig:excav_perf}.2, the ACR method successfully displaced and reoriented the faulty robot into a less obstructive position in two of the three trials. The third trial is not considered a failure, as no total obstruction occurred. Typically, total obstruction arises when the faulty robot is oriented perpendicular to the tunnel, preventing the active robots from bypassing it. In contrast, the Baseline method resulted in more obstructive final configurations across all trials, which directly contributed to the reduced excavation performance (see \href{https://youtu.be/tZFPhAJ8bNo}{youtube link} for demo). 

Figure~\ref{fig:robot_trajectory} illustrates the robot trajectories in one trial for each method. The Baseline method trajectories depict a scenario where active robots unintentionally pushed the faulty robot deeper into the tunnel, exacerbating the obstruction near the excavation site. In contrast, the ACR method resulted in active robots consistently pushing the faulty robot out of the tunnel towards the home area, reducing its obstruction potential. As previously discussed, the orientation of the faulty robot plays an important role in determining the level of obstruction. Faulty robots positioned perpendicularly to the tunnel create the most severe obstructions, which was more common in the baseline trials.

Contact interactions in the ACR method play a dual role: first, they enable robots to learn the collision map to infer the dynamics of the tunnel; second, they help regulate the number of active robots based on the parameter ($P_e$) ~\cite{frontiers_kaina}. The position of the faulty robot significantly impacts collision frequency, which, in turn, influences the workload distribution. When the faulty robot is located at the tunnel’s extremities (e.g., near the home area or close and parallel to the wall), it causes minimal collision, thereby maximizing the number of active robots and reducing the need for ACR activation. However, when the faulty robot is positioned in the middle of the tunnel or near the excavation area, it creates a substantial bottleneck, decreasing the number of active robots and increasing the likelihood of ACR activation. This adaptability indicates the robustness of the ACR method across different initial configuration of the faulty robot.

Despite the robustness of the ACR method, certain limitations remain. Specifically, the method struggles with edge cases where the faulty robot is positioned in dead-end areas (e.g., excavation zone), rendering active pushing ineffective. Additionally, as the number of robots increases, convergence becomes more difficult due to overcrowding and the highly non-stationary environment, which impedes task execution. To mitigate this challenge, a ``curriculum learning'' approach could be employed, where the the number of robots is increased gradually. 

Comparing our method to existing fault-tolerance techniques presents challenges due to the unique reliance of the ACR method on contact-based interactions. Most contemporary methods depend on precise state estimation, localization, and inter-robot communication for centralized control. By contrast, the ACR method is designed for extreme environments with limited resources, making it highly suited for such conditions. A statistical analysis was conducted on three trials per method, with ANOVA tests performed on the results (Baseline: one-way ANOVA $F=0.74$, $P=0.49$; ACR: one-way ANOVA $F=0.31$, $P=0.74$). The $p-values$ exceeded the significance threshold of $\alpha = 0.05$, confirming that the trials adequately capture the distribution of experimental data for our objectives.

\section{CONCLUSION}

In this work, we demonstrated how leveraging contact interactions can enable emergent fault tolerance in decentralized multi-robot systems, even in environments where precise sensing and global control mechanisms are absent. By utilizing noisy contact measurements and localization, each robot is able to estimate the conditional likelihood of encountering a faulty robot. This allows individual robots to adjust their behavior to collectively reposition the faulty robot into a configuration that minimizes interference and maintains optimal group performance. The proposed Active Contact Response (ACR) algorithm enables robots to autonomously recover from system disruptions caused by individual robot failures, ensuring continuity in collective tasks.

The ACR technique has the potential to be scaled to larger teams of robots, offering promise for broader applications. Future research will investigate the scalability and generalizability of the ACR method, particularly by simulating larger multi-robot systems and comparing the performance of ACR against other existing fault-tolerance techniques.
In more complex, dynamic environments with a higher number of robots, introducing local information exchange—such as collision map sharing—could further enhance system performance. Additionally, this technique has the potential to be applied in other multi-agent scenarios, and  
smart active matter systems, where individual agents can leverage social, local, and physical interactions to improve fault tolerance and tackle more intricate challenges. This work opens avenues for expanding fault-tolerant strategies in distributed, resource-constrained robotic systems.

\section*{ACKNOWLEDGMENT}

We are grateful to Michael A. D. Goodisman, and Ram Avinery of Georgia Institute of Technology for their helpful discussions and suggestions. This research was funded by ARO MURI award W911NF-19-1-023 and NSF grant IOS-2019799.

%
%

\bibliographystyle{unsrt} 
\bibliography{bibliography}

\begin{thebibliography}{10}

\bibitem{lesser1999multi}
Victor Lesser, Michael Atighetchi, Brett Benyo, Bryan Horling, Anita Raja, Regis Vincent, Thomas Wagner, Ping Xuan, and SX~Zhang.
\newblock A multi-agent system for intelligent environment control.
\newblock In {\em Proceedings of the Third International Conference on Autonomous Agents}, volume~10, pages 301136--301213. ACM Press New York, NY, 1999.

\bibitem{chen2009integrating}
Bo~Chen, Harry~H Cheng, and Joe Palen.
\newblock Integrating mobile agent technology with multi-agent systems for distributed traffic detection and management systems.
\newblock {\em Transportation Research Part C: Emerging Technologies}, 17(1):1--10, 2009.

\bibitem{drew2021multi}
Daniel~S Drew.
\newblock Multi-agent systems for search and rescue applications.
\newblock {\em Current Robotics Reports}, 2:189--200, 2021.

\bibitem{ROUSSET201627}
Alban Rousset, Bénédicte Herrmann, Christophe Lang, and Laurent Philippe.
\newblock A survey on parallel and distributed multi-agent systems for high performance computing simulations.
\newblock {\em Computer Science Review}, 22:27--46, 2016.

\bibitem{valckenaers2007applications}
Paul Valckenaers, John Sauter, Carles Sierra, and Juan~Antonio Rodriguez-Aguilar.
\newblock Applications and environments for multi-agent systems.
\newblock {\em Autonomous Agents and Multi-Agent Systems}, 14:61--85, 2007.

\bibitem{baudoin2010using}
Y~Baudoin and Maki~K Habib.
\newblock {\em Using robots in hazardous environments: Landmine detection, de-mining and other applications}.
\newblock Elsevier, 2010.

\bibitem{shakshuki2015multi}
Elhadi Shakshuki and Malcolm Reid.
\newblock Multi-agent system applications in healthcare: current technology and future roadmap.
\newblock {\em Procedia Computer Science}, 52:252--261, 2015.

\bibitem{sidd_inproceedings}
Siddharth Mayya, Pietro Pierpaoli, Girish Nair, and Magnus Egerstedt.
\newblock Collisions as information sources in densely packed multi-robot systems under mean-field approximations.
\newblock 07 2017.

\bibitem{article_Schmickl}
Thomas Schmickl, Ronald Thenius, Christoph Möslinger, Gerald Radspieler, Serge Kernbach, Marc Szymanski, and Karl Crailsheim.
\newblock Get in touch-cooperative decision making based on robot-to-robot collisions.
\newblock {\em Autonomous Agents and Multi-Agent Systems}, 18:133--155, 02 2009.

\bibitem{Slavkov2018}
I.~Slavkov, D.~Carrillo-Zapata, N.~Carranza, X.~Diego, F.~Jansson, J.~Kaandorp, S.~Hauert, and J.~Sharpe.
\newblock Morphogenesis in robot swarms.
\newblock {\em Science Robotics}, 3(25):eaau9178, 2018.

\bibitem{article_Siddharth_and_Wilson}
Siddharth Mayya, Sean Wilson, and Magnus Egerstedt.
\newblock Closed-loop task allocation in robot swarms using inter-robot encounters.
\newblock {\em Swarm Intelligence}, 13:115--143, 06 2019.

\bibitem{8794124}
Siddharth Mayya, Pietro Pierpaoli, and Magnus Egerstedt.
\newblock Voluntary retreat for decentralized interference reduction in robot swarms.
\newblock In {\em 2019 International Conference on Robotics and Automation (ICRA)}, pages 9667--9673, 2019.

\bibitem{Gravish2012}
Nick Gravish, Mateo Garcia, Nicole Mazouchova, Laura Levy, Paul~B. Umbanhowar, Michael A.~D. Goodisman, and Daniel~I. Goldman.
\newblock {Effects of worker size on the dynamics of fire ant tunnel construction}.
\newblock {\em Journal of The Royal Society Interface}, 9(77):3312--3322, dec 2012.

\bibitem{gordon1996organization}
Deborah~M Gordon.
\newblock The organization of work in social insect colonies.
\newblock {\em Nature}, 380(6570):121--124, 1996.

\bibitem{osti_10427980}
Ram Avinery, Kehinde~O. Aina, Carl~J. Dyson, Hui-Shun Kuan, Meredith~D. Betterton, Michael~A. Goodisman, and Daniel~I. Goldman.
\newblock Agitated ants: regulation and self-organization of incipient nest excavation via collisional cues.
\newblock {\em Journal of The Royal Society Interface}, 20(202).

\bibitem{monaenkova2015behavioral}
Daria Monaenkova, Nick Gravish, Greggory Rodriguez, Rachel Kutner, Michael~AD Goodisman, and Daniel~I Goldman.
\newblock Behavioral and mechanical determinants of collective subsurface nest excavation.
\newblock {\em The Journal of experimental biology}, 218(9):1295--1305, 2015.

\bibitem{Gravish2015}
Nick Gravish, Gregory Gold, Andrew Zangwill, Michael A.~D. Goodisman, and Daniel~I. Goldman.
\newblock {Glass-like dynamics in confined and congested ant traffic}.
\newblock {\em Soft Matter}, 11(33):6552--6561, 2015.

\bibitem{Gravish9746}
Nick Gravish, Daria Monaenkova, Michael A.~D. Goodisman, and Daniel~I. Goldman.
\newblock Climbing, falling, and jamming during ant locomotion in confined environments.
\newblock {\em Proceedings of the National Academy of Sciences}, 110(24):9746--9751, 2013.

\bibitem{Souliman2014}
Aya Souliman, Abdulkader Joukhadar, Hamid Alturbeh, and James F.~Whidborne.
\newblock {\em Intelligent Collision Avoidance for Multi Agent Mobile Robots}, pages 297--315.
\newblock Springer International Publishing, Cham, 2014.

\bibitem{HernandezMartinez2011}
Eduardo~G. Hern{\'{a}}ndez-Mart{\'{i}}nez and Eduardo Aranda-Bricaire.
\newblock {Convergence and Collision Avoidance in Formation Control: A Survey of the Artificial Potential Functions Approach}.
\newblock In {\em Multi-Agent Systems - Modeling, Control, Programming, Simulations and Applications}, number 1997. InTech, apr 2011.

\bibitem{9123564}
Jinwen Hu, Houxin Zhang, Lu~Liu, Xiaoping Zhu, Chunhui Zhao, and Quan Pan.
\newblock Convergent multiagent formation control with collision avoidance.
\newblock {\em IEEE Transactions on Robotics}, 36(6):1805--1818, 2020.

\bibitem{Arslan2016}
Omur Arslan, Dan~P. Guralnik, and Daniel~E. Koditschek.
\newblock {Coordinated Robot Navigation via Hierarchical Clustering}.
\newblock {\em IEEE Transactions on Robotics}, 32(2):352--371, apr 2016.

\bibitem{Roy2019}
DIbyendu Roy, Arijit Chowdhury, Madhubanti Maitra, and Samar Bhattacharya.
\newblock {Virtual Region based Multi-robot Path Planning in an Unknown Occluded Environment}.
\newblock In {\em 2019 IEEE/RSJ International Conference on Intelligent Robots and Systems (IROS)}, pages 588--595. IEEE, nov 2019.

\bibitem{TAYLOR2021103754}
Chris Taylor and Cameron Nowzari.
\newblock The impact of catastrophic collisions and collision avoidance on a swarming behavior.
\newblock {\em Robotics and Autonomous Systems}, 140:103754, 2021.

\bibitem{BORRMANN201568}
Urs Borrmann, Li~Wang, Aaron~D. Ames, and Magnus Egerstedt.
\newblock Control barrier certificates for safe swarm behavior.
\newblock {\em IFAC-PapersOnLine}, 48(27):68--73, 2015.
\newblock Analysis and Design of Hybrid Systems ADHS.

\bibitem{Siddharth_1109}
Siddharth Mayya, Pietro Pierpaoli, Girish Nair, and Magnus Egerstedt.
\newblock Localization in densely packed swarms using interrobot collisions as a sensing modality.
\newblock {\em Trans. Rob.}, 35(1):21–34, February 2019.

\bibitem{frontiers_kaina}
Kehinde~O. Aina, Ram Avinery, Hui-Shun Kuan, Meredith~D. Betterton, Michael A.~D. Goodisman, and Daniel~I. Goldman.
\newblock Toward task capable active matter: Learning to avoid clogging in confined collectives via collisions.
\newblock {\em Frontiers in Physics}, 10, 2022.

\bibitem{science_paper}
J.~Aguilar, D.~Monaenkova, V.~Linevich, W.~Savoie, B.~Dutta, H.-S. Kuan, M.~D. Betterton, M.~A.~D. Goodisman, and D.~I. Goldman.
\newblock Collective clog control: Optimizing traffic flow in confined biological and robophysical excavation.
\newblock {\em Science}, 361(6403):672--677, 2018.

\bibitem{5229934}
Gottfried Koppensteiner, Munir Merdan, Wilfried Lepuschitz, and Ingo Hegny.
\newblock Hybrid based approach for fault tolrance in a multi-agent system.
\newblock In {\em 2009 IEEE/ASME International Conference on Advanced Intelligent Mechatronics}, pages 679--684, 2009.

\bibitem{inproceedings_Almeida}
De~Almeida, Jean-Pierre Briot, Samir Aknine, Zahia Guessoum, and Olivier Marin.
\newblock Towards autonomic fault-tolerant multi-agent systems.
\newblock 01 2007.

\bibitem{Kumar_Sanjeev}
Sanjeev Kumar and Philip~R. Cohen.
\newblock Towards a fault-tolerant multi-agent system architecture.
\newblock In {\em Proceedings of the Fourth International Conference on Autonomous Agents}, AGENTS '00, page 459–466, New York, NY, USA, 2000. Association for Computing Machinery.

\bibitem{fedoruk2002improving}
Alan Fedoruk and Ralph Deters.
\newblock Improving fault-tolerance by replicating agents.
\newblock In {\em Proceedings of the first international joint conference on Autonomous agents and multiagent systems: part 2}, pages 737--744, 2002.

\bibitem{Marin2001TowardsAF}
Olivier Marin, Pierre Sens, Jean-Pierre Briot, and Zahia Guessoum.
\newblock Towards adaptive fault-tolerance for distributed multi-agent systems.
\newblock 2001.

\bibitem{guessoum2005adaptive}
Zahia Guessoum, Nora Faci, and Jean-Pierre Briot.
\newblock Adaptive replication of large-scale multi-agent systems: towards a fault-tolerant multi-agent platform.
\newblock {\em ACM SIGSOFT Software Engineering Notes}, 30(4):1--6, 2005.

\bibitem{1639670}
Ad.L. Almeida, S.~Aknine, J.-P. Briot, and J.~Malenfant.
\newblock Plan-based replication for fault-tolerant multi-agent systems.
\newblock In {\em Proceedings 20th IEEE International Parallel \& Distributed Processing Symposium}, pages 7 pp.--, 2006.

\bibitem{5952430}
Aarti Singh, Dimple Juneja, and A.K. Sharma.
\newblock Adaptive and automated fault-tolerance for multi-agent systems.
\newblock In {\em 2011 IEEE International Conference on Computer Science and Automation Engineering}, volume~1, pages 53--57, 2011.

\bibitem{Faci_fault_tolerant_mas}
Nora Faci, Zahia Guessoum, and Olivier Marin.
\newblock Dimax: A fault-tolerant multi-agent platform.
\newblock In {\em Proceedings of the 2006 International Workshop on Software Engineering for Large-Scale Multi-Agent Systems}, SELMAS '06, page 13–20, New York, NY, USA, 2006. Association for Computing Machinery.

\bibitem{FaultToleranceMAS}
Samuel~H. Christie~V and Amit~K. Chopra.
\newblock Fault tolerance in multiagent systems.
\newblock In Cristina Baroglio, Jomi~F. Hubner, and Michael Winikoff, editors, {\em Engineering Multi-Agent Systems}, pages 78--86, Cham, 2020. Springer International Publishing.

\bibitem{7487153}
Hyongju Park and Seth Hutchinson.
\newblock An efficient algorithm for fault-tolerant rendezvous of multi-robot systems with controllable sensing range.
\newblock In {\em 2016 IEEE International Conference on Robotics and Automation (ICRA)}, pages 358--365, 2016.

\bibitem{7353788}
Hyongju Park and Seth Hutchinson.
\newblock A distributed robust convergence algorithm for multi-robot systems in the presence of faulty robots.
\newblock In {\em 2015 IEEE/RSJ International Conference on Intelligent Robots and Systems (IROS)}, pages 2980--2985, 2015.

\bibitem{7535869}
Joelle Al~Hage, Maan~E. El~Najjar, and Denis Pomorski.
\newblock Fault tolerant collaborative localization for multi-robot system.
\newblock In {\em 2016 24th Mediterranean Conference on Control and Automation (MED)}, pages 907--913, 2016.

\bibitem{8854221}
Hao Yang, Qing-Long Han, Xiaohua Ge, Lei Ding, Yuhang Xu, Bin Jiang, and Donghua Zhou.
\newblock Fault-tolerant cooperative control of multiagent systems: A survey of trends and methodologies.
\newblock {\em IEEE Transactions on Industrial Informatics}, 16(1):4--17, 2020.

\bibitem{aguilar2018collective}
J~Aguilar, D~Monaenkova, V~Linevich, W~Savoie, B~Dutta, H-S Kuan, MD~Betterton, MAD Goodisman, and DI~Goldman.
\newblock Collective clog control: Optimizing traffic flow in confined biological and robophysical excavation.
\newblock {\em Science}, 361(6403):672--677, 2018.

\bibitem{PANESCU2007121}
Doru Panescu and ştefan Dumbrava.
\newblock Multi-robot systems: From finite automata to multi-agent systems.
\newblock {\em IFAC Proceedings Volumes}, 40(18):121--126, 2007.
\newblock 4th IFAC Conference on Management and Control of Production and Logistics.

\end{thebibliography}



\end{document}